\documentclass[11pt,a4paper]{article}
\usepackage[hyperref]{acl2020}
\usepackage{times}
\usepackage{latexsym}
\usepackage{graphicx}
\usepackage{multirow}
\usepackage{array}
\usepackage{hhline}
\usepackage[inline]{enumitem}

\usepackage{arydshln}
\usepackage{color}
\usepackage{soul}

\newcolumntype{L}{>{\arraybackslash}m{8cm}}

\usepackage{microtype}

\aclfinalcopy 

\title{Neural Passage Retrieval with Improved Negative Contrast}

\author{Jing Lu$^1$\thanks{~~Work done during an internship at Google.}, Gustavo Hern\'andez \'Abrego$^2$, Ji Ma$^2$, Jianmo Ni$^2$, Yinfei Yang$^2$ \\
  $^1$Human Language Technology Research Institute, University of Texas at Dallas \\
  $^2$Google Research \\
  \texttt{jing.lu@utdallas.edu} \\
  \texttt{$\{$gustavoha, maji, jianmon, yinfeiy$\}$@google.com} \\}

\date{}

\begin{document}
\maketitle

\begin{abstract}
    In this paper we explore the effects of negative sampling in dual encoder models used to retrieve passages for automatic question answering.
    We explore four negative sampling strategies that complement the straightforward random sampling of negatives, typically used to train dual encoder models.
    Out of the four strategies, three are based on retrieval and one on heuristics.
    Our retrieval-based strategies are based on the semantic similarity and the lexical overlap between questions and passages.
    We train the dual encoder models in two stages: pre-training with synthetic data and fine tuning with domain-specific data.
    We apply negative sampling to both stages.
    The approach is evaluated in two passage retrieval tasks.
    Even though it is not evident that there is one single sampling strategy that works best in all the tasks, it is clear that our strategies contribute to improving the contrast between the response and all the other passages.
    Furthermore, mixing the negatives from different strategies achieve performance on par with the best performing strategy in all tasks. 
    Our results establish a new state-of-the-art level of performance on two of the open-domain question answering datasets that we evaluated.
\end{abstract}
\section{Introduction}

Automatic question answering is a very active area of research within natural language processing.
One possible way to approach this task is to look for answers in the text passages of a collection of documents.
Recent research has shown promising results~\cite{use-qa,multireqa,dpr,ance} on developing neural models for passage retrieval tasks, including Retrieval Question Answering~\cite{reqa}, Open Domain Question Answering~\cite{Chen_2017}, and MS MARCO~\cite{msmarco}.
The models in these systems are often trained using the dual encoder framework~\cite{yang-etal-2018-learning,gillick2018endtoend} where questions and passages are encoded separately.
Training an effective neural retrieval model usually requires a large amount of high-quality data.
To alleviate the need of high-quality data, training can be approached in two-stages: pre-training on noise data~\cite{guo-etal-2018-effective,Chang2020Pre-training,ma2020zeroshot} and fine tuning on a smaller amount of high-quality data, also regarded as ``gold" data.

\begin{figure}[t]
    \centering
    \includegraphics[width=\linewidth]{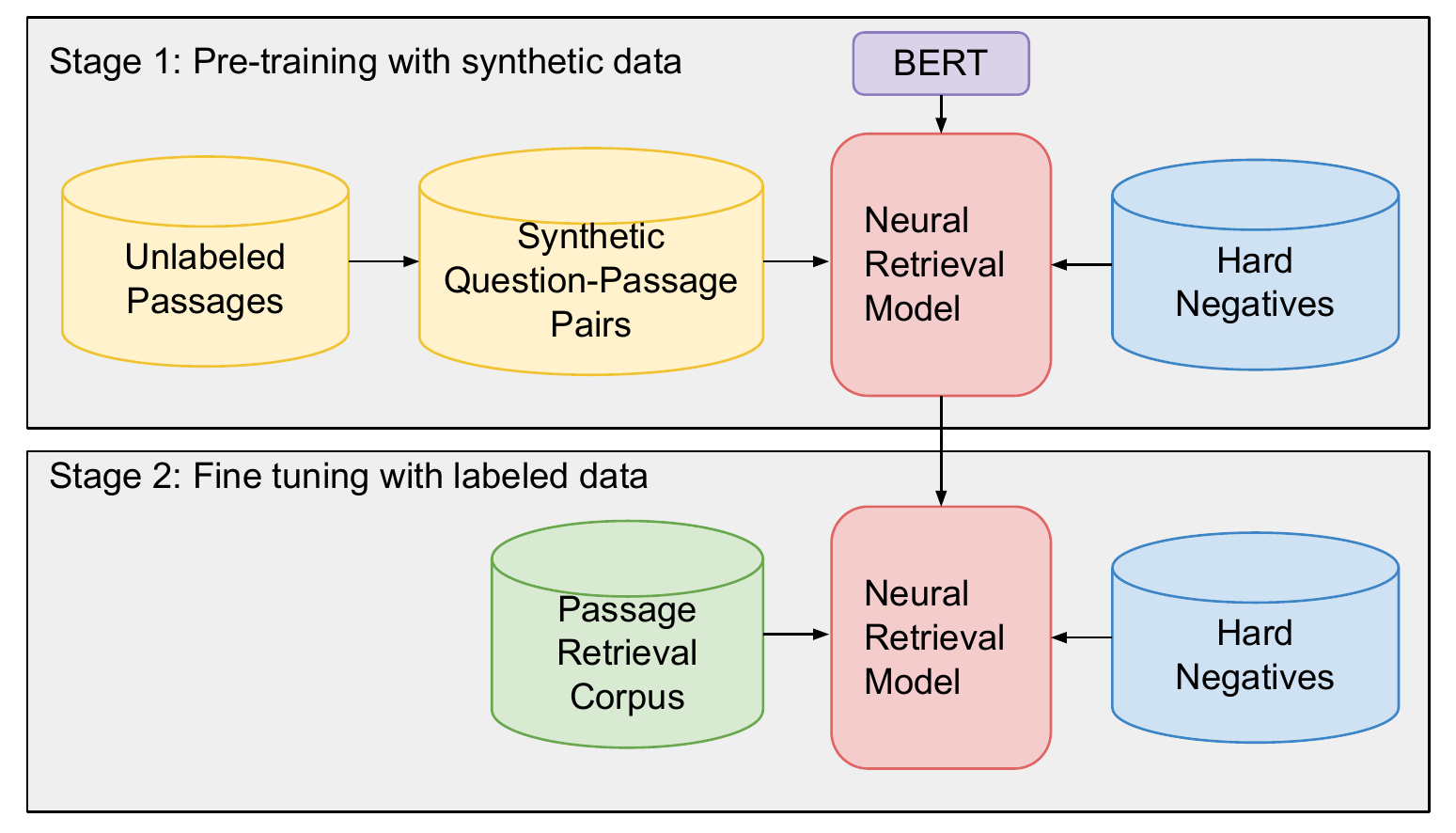}
    \caption{Two-stages neural retrieval model with hard negatives in both stages. In Stage 1, the model is trained using synthetic question-passages pairs. In Stage 2, the model is fine tuned using gold training data.}
    \label{fig:pipeline}
\end{figure}

When used for question answering, one advantage of the dual encoder is that training in batches allows to use, for each question, the passages that answer all the other questions in the batch as negatives~\cite{gillick2018endtoend}.
Given that the training batches are randomly sampled from all the question-passage pairs, the negatives in the batch are random in nature.
While effective in many retrieval tasks~\cite{henderson2017efficient,yang-etal-2018-learning}, random negatives have the limitation of not being targeted nor challenging enough to clearly separate the passage that answers a given question from any other passage.
How to sample the negatives in a way that widens this separation and improves the contrast between the correct and incorrect passages remains an open question.

In this paper we systematically explore the use of ``hard'' negatives in the neural passage retrieval models that we train using a two-stage approach. Using hard negatives as part of the dual encoder framework has shown advantageous in different tasks \cite{guo-etal-2018-effective,gillick-etal-2019-learning,dpr, ance}.
We explore different types of negatives, and experiment using them in both the pre-training and fine-tuning stages.
The types of negatives we tried are:
\begin{enumerate*}[label=({\arabic*})]
\item[1.] \textit{Coarse semantic similarity retrieval-based negatives;}
\item[2.] \textit{Fine semantic similarity retrieval-based negatives\footnote{For details on the definition of ``coarse" and ``fine", refer to section \ref{sec:negatives} below};}
\item[3.] \textit{BM25 retrieval-based negatives;}
\item[4.] \textit{Heuristics-based context negatives.}
\end{enumerate*}

We first use hard negatives on the data that we use to pre-train the models.
We leverage the question generator model described in~\cite{ma2020zeroshot} and generate new questions for each of the passages we use in the pre-training stage (Stage 1).
During pre-training we use negatives generated from strategy 4\footnote{Or strategy 1, if strategy 4 is not feasible} to improve the retrieval model,
as the other strategies could introduce more false negatives into the data.
Next, we continue with the fine tuning stage (Stage 2) using a small amount of gold training data.
At this stage, we explore all four types of negative sampling.
To the best of our knowledge, this is the first work that explores the effectiveness of hard negatives for passage retrieval in a systematic way, and integrates them in the retrieval models  pre-training stage.
Our overall experimental architecture is outlined in Figure~\ref{fig:pipeline}.


We conduct experiments with this approach on two passage retrieval tasks: Open Domain QA (using Natural Question~(NQ) and SQuAD) and MS MARCO~\cite{msmarco}.
Our results show that all four kinds of hard negatives improve the dual encoder models significantly with consistent performance gains across both tasks.
However, depending on the types of questions and their domain, one kind of hard negative may perform better than the others in a particular task.
For example, context negatives work best in NQ and semantic retrieval-based negatives (both coarse and fine) work best in SQuAD.
We further ensemble the models trained on different types of hard negatives.
The final models achieve state-of-the-art performance on Open Domain QA task with an improvement over prior works of 0.8--2.9 points on accuracy rates. 

The main contribution of this paper are:
\begin{enumerate*}[label=({\arabic*})]
\item Systematically explore hard negatives as a way to augment the contrast between the target passage and any other passage; 
\item A novel pre-training approach that integrates synthetic question generation with hard negatives;
\item An ensemble approach combining models trained from different hard negatives that establishes a new state-of-the-art performance level in multiple passage-retrieval tasks.
\end{enumerate*}

\section{Related Work}

Recent advances in deep-learning have spurred a surge of interests in adopting neural models to information retrieval.
Many existing works adopt a pipeline approach that includes a retrieval stage and a reranking stage.
For first-stage retrieval, where the task is to find the most relevant documents in a large document collection, 
one line of research investigates projecting query and documents into a shared dense space \cite{palangi2016deep, gillick2018endtoend, dpr, luan2020sparse}.
The models in this family are regarded as dense-retrieval models or dual encoder models.
This is the kind of model that we employ in this work.
Ideally, a query and its relevant documents should be projected in each other's vicinity.
From this point of view, finding the most relevant documents can be cast to a nearest neighbor search \cite{liu2011hashing,JDH17}.   

Previous attempts at improving the quality of dual encoder models can be classified into two types.
The first type focuses on finding a good initialization for the model parameters.
This is typically achieved by pre-training the model on various tasks \cite{bert, lee2019latent, Chang2020Pre-training}.  
Another type of approach focuses on learning better representations using hard negatives.
This strategy has proved to be effective in passage retrieval tasks \cite{dpr,ance}, machine translation \cite{guo-etal-2018-effective} and entity linking \cite{gillick-etal-2019-learning}.
These works mine hard negatives using different strategies.
For example, \newcite{guo-etal-2018-effective} mine ``coarse'' negatives with a low-resolution model.
\newcite{dpr} mine hard negatives using a BM25 model, and \newcite{gillick-etal-2019-learning} use a model trained with random negatives and select examples that are ranked above the correct one as negative examples.

Neural networks have also become popular at the reranking stage~\cite{Guo2016ADR, DeepRelevenceRanking, nogueira2019passage, macavaney2019cedr}. 
This paper focus on the retrieval stage, and leaves reranking as a future line of work.

\section{Neural Passage Retrieval Model}
\label{sec:model}

\begin{figure}[t]
    \centering
    \includegraphics[width=5cm]{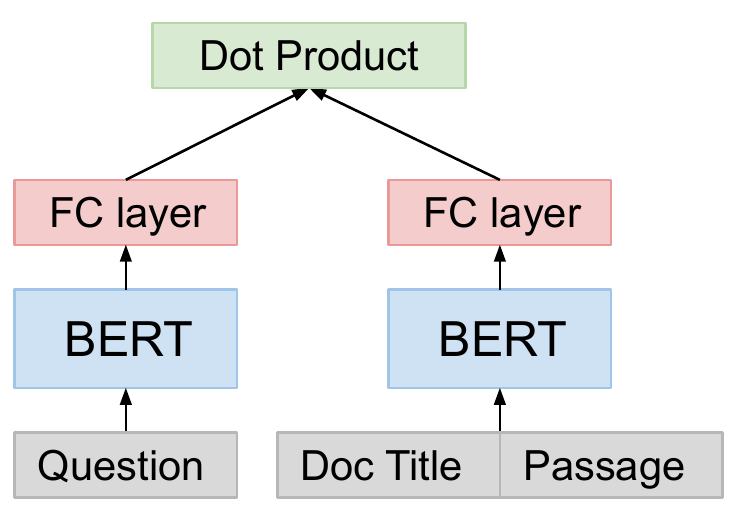}
    \caption{The neural passage retrieval model. The document title and passage are concatenated and fed into the passage encoder.}
    \label{fig:model}
\end{figure}

In this paper, the backbone of our experimental setup is a BERT-based dual encoder, similar to the one used in the Dense Passage Retrieval~(\textbf{DPR}) work by \citet{dpr}.
The architecture of our model is shown in Figure~\ref{fig:model}.
For a given question-passage pair, the question text is fed through the left side into a BERT transformer. 
The hidden vector of the [CLS] token is then fed into a fully-connected linear layer (FC layer) of dimension 768. 
The output of the FC layer is used as the representation of the question. 
On the right side, the title of the document from where the passage was extracted is concatenated with the text of the passage.
The concatenation is represented as: [CLS] \textit{title} [SEP] \textit{passage} [SEP] and it is fed into the same BERT transformer. 
Similarly, the hidden vector of the [CLS] token is then fed into the same FC layer and the resulting output vector is used as the passage representation. 
Note that, in contrast to the standard BERT pooled output, we take the hidden vector of [CLS] directly without having an additional layer with a $tanh$ activation function.
The final question and passage embeddings are then $l_2$ normalized. 
In our experiments we used the uncased $BERT_{base}$ model.

During training, passage retrieval is modeled as a ranking problem.
We use a bidirectional learning objective to optimize the ranking of the retrieved passages in both the forward and backward directions.
For the forward ranking, we rank $y_i$, the relevant passage for question $x_i$, over all other passages in the same batch. 
For the backward ranking, we rank $x_i$ over all other questions in the same batch.
We use the dot-product as scoring function to measure the similarity between questions and passages.
The loss function is the unweighted average of both the forward-ranking $L_f$ and backward-ranking $L_b$ losses:

\begin{small}
\begin{equation}
L_f = -\frac{1}{N} \sum_{i=1}^N log \frac{e^{\phi(x_i, y_i)}}{e^{\phi(x_i, y_i)} + \sum_{n=1, n \neq i}^N e^{\phi(x_i, y_n)}}
\end{equation}
\end{small}
\vspace{-2mm}
\begin{small}
\begin{equation}
L_b = -\frac{1}{N} \sum_{i=1}^N log \frac{e^{\phi(y_i, x_i)}}{e^{\phi(y_i, x_i)} + \sum_{n=1, n \neq i}^N e^{\phi(y_i, x_n)}}
\end{equation}
\end{small}
\vspace{-2mm}

\begin{small}
\begin{equation}
L = 0.5 \times (L_f + L_b)
\end{equation}
\end{small}

During inference, we encode all the passages from our document collection using the passage encoder and index them using ScaNN \cite{guo2019accelerating}. 
ScaNN performs efficient dense vector similarity search as other open-sourced tools such as FAISS\cite{JDH17}, SPTAG \cite{ChenW18} do.
In such a way, given a question encoding, we retrieve from the index the $k$ passages closest to the question.


\section{Improved Negative Contrast}
\label{sec:negatives}

We first describe the negative sampling strategies used in pre-training (Stage 1) and fine-tuning (Stage 2).
Then, we describe how the hard negatives are used to train the neural retrieval models.

\subsection{Hard Negatives in Pre-training}

In this work, we adopt the approach mentioned in \cite{ma2020zeroshot}, and trained our own question generation model based on Text-To-Text Transfer Transformer (T5) \cite{t5}. 
We then use this model to generate synthetic questions for the passages in a collection of documents. 
Thus, we train our first-stage retrieval model using millions of synthetic question-passage pairs. 

As the generated question-passage pairs can be noisy, retrieval-based negatives using BM25 or a semantic similarity model could end up generating negative pairs that are better (less noisy) than the ``positive pairs'' resulting from the question generation.
To avoid this undesirable condition, we use a heuristic-based definition of hard negatives at this stage, if possible.
Given a question and its corresponding passage, we define the \textbf{context negatives} as the passages from the same document that are not the corresponding passage. 
Intuitively, other passages (even if they come from the same document) should not have higher relevance to the question than the passage which was used to generate the question.
If the document contains only one passage, we split the passage in half and pick as the context negative the half that does not contain the span of text that answers the question.
However, this heuristics assumes that there is a mapping between documents and passages.
That may not always be the case, as described in Section \ref{sec:msmarcoexperiments} regarding one of our testing tasks.

\subsection{Hard Negatives in Fine Tuning}

At Stage 2 of the training we are able to explore types of hard negatives that are defined in terms of their distance to the gold passages. 
When retrieving cross-lingual text, \newcite{guo-etal-2018-effective} showed that hard negatives that are merely topically related to the actual translations, or that are semantically similar but sufficiently different, could hinder the retrieval performance.
In contrast, hard negatives that are semantically similar to the translation but are not its identical improve the performance. 
In this paper, we systematically study the effects of the following kinds of hard negatives for passage retrieval.

\textbf{Retrieval negatives -- coarse semantic similarity.} As in \cite{guo-etal-2018-effective}, these are negatives that are on-topic but the semantic similarity between them and the answer passage is not that close. 
To retrieve the coarse hard negatives, we separately train a low-dimensional dual encoder model on the question-passage pairs.
Experimentally we have found that embeddings of dimension size 25 is sufficient to produce this kind of negatives.
We use the low-dimensional embeddings to perform an approximate search of nearest neighbors from all passages in the document collection for each question in the training set.
The top $M$ passages with embeddings closest (in terms of their dot-product) to the question are retrieved as coarse hard negatives.

\textbf{Retrieval negatives -- fine semantic similarity.} 
These are passages that are not only on-topic but also semantically similar to the answer passage. 
To retrieve these negatives, we train a high-dimensional dual encoder model on the question-passage pairs.
Experimentally we have found that embeddings from a model of dimension size 512 do a good job retrieving semantically similar passages. 
We follow the same approach as with the coarse negatives to get $M$ passages for each question in the training data as the fine hard negatives.  

\textbf{Retrieval negatives -- BM25.}
This type of hard negatives showed to be effective in the work of \newcite{dpr}.
Even though in this work we focus on dense retrieval models, sparse models such as BM25 still provide strong performance.
Especially when the tokens in the question and the tokens in the answer passage greatly overlap. 
To retrieve this kind of hard negatives we collect the top passages returned by BM25 and filter out passages that contain the answer spans.

\textbf{Context negatives.} 
Just as we did in Stage 1, we also employ the context negatives in Stage 2 and follow the same strategy to generate them.

To illustrate our  negative sampling strategies, examples of the four hard negative types are shown and discussed in section \ref{sec:example}.

\subsection{Training with Hard Negatives}
\label{sec:hnintraining}
For each question, and assuming we have extracted $M$ hard negatives for each of them, our procedure to train the neural retrieval model with hard negatives is to use them in addition to the random negatives sampled from the same batch.
In each training iteration, we randomly select $N$ negatives from the pool of $M$ hard negatives for each question.
The hard negatives are appended to the random negatives in the standard dual encoder training\footnote{The hard negatives are only applied to the question to passage training loss during training but not vice versa.}.
Note that the hard negatives for one question are treated as random negatives for the other questions in the same batch.
Therefore, for a batch of size $B$, each question is compared during training against $(N+1) \times B$ passages instead of just $B$ passages in the standard way to train a dual encoder. 

\subsection{Ensemble}
We study two ensemble methods to investigate how models trained with different hard negatives could complement each other. 

\textbf{Embedding fusion.} The question (or document) embeddings obtained from each model are multiplied by different coefficients and then concatenated as the ensemble embeddings. The coefficients are tuned based on the performance on the development set. Then the ensemble embeddings are used for retrieving the corresponding passages for the questions. The advantage of this fusion is that we only need to perform the retrieval once.

\textbf{Rank fusion.} Following the Reciprocal Rank Fusion (RRF) \cite{rrf} method, the final ranking results are obtained by considering the ranking positions of each candidate in the rankings generated by different models.

\section{Experimental Setup}
We evaluate our proposed model on two tasks: firstly, we evaluate on the passage retrieval task for open-domain question answering with the goal of retrieving passages that contain the correct answer spans given a question. 
Secondly, to understand how our model perform on large-scale text retrieval datasets, we also evaluate on the MS MARCO passage ranking task. 

\subsection{Open-Domain QA Retrieval}
We evaluate on two open-domain QA datasets: Natural Question (\textbf{NQ})\cite{nq} and \textbf{SQuAD} \cite{squad}.
NQ contains questions from real Google search queries and the answers from Wikipedia articles identified by annotators. 
We follow \newcite{lee2019latent} and convert the dataset to a format suitable for open-domain QA.
Specifically, we only keep questions with short answers (no more than five tokens).
On the other hand, SQuAD v1.1 is a commonly used dataset for reading comprehension tasks.
In contrast to NQ, the questions in SQuAD are generated by annotators given paragraphs from Wikipedia.
The number of questions in each dataset is shown in Table~\ref{tab:dataset}.

In our experiments we use Wikipedia as our collection of documents and knowledge source from where to retrieve passages that answer the questions.
Following \newcite{lee2019latent} and \newcite{dpr}, we use an English Wikipedia dump from Dec.20, 2018.
After filtering semi-structured data, such as tables and info-boxes, each document is split into disjoint text passages of 100 words.
In such a way, we get 21,015,324 passages in total.
In order to be able to compare our work with the models from \newcite{dpr} directly, we use the preprocessed Wikipedia passages as released by the authors\footnote{https://github.com/facebookresearch/DPR}.

To generate synthetic questions, the question generation model is trained on \textbf{NQ}. 
We then apply the model on each passage in the Wikipedia corpus.
Specifically, we generate three questions per passage, which results in 62 million synthetic question-passage pairs in total\footnote{Examples of synthetic questions can be found in the Appendix.}. 

We report the results using Top-K accuracy, which is the fraction of $K$ retrieved passages that contain a span with the answer to the question. 
Specifically, we report results for $K$ = [1, 5, 10, 20, 100]

\begin{table}[th!]
\centering
\begin{tabular}{ lrrr } 
 \hline
 \textbf{Dataset} & \textbf{Train} & \textbf{Dev} & \textbf{Test} \\ \hline
 NQ &  58,880 & 6,515 & 3,610 \\ 
 SQuAD v1.1 &  70,096 & 7,921 & 10,570 \\ 
 MS MARCO & 532,761 & 6,980 &  200 \\ \hline
 \end{tabular}
 \caption{Number of examples in Train/Dev/Test sets}
 \label{tab:dataset}
 \end{table}

\subsection{MS MARCO Passage Ranking}
The MS MARCO passage ranking task\footnote{https://github.com/microsoft/MSMARCO-Passage-Ranking} consists of two sub-tasks: a full retrieval and a top-1000 reranking task.
In this paper we evaluate on the full retrieval task only.
The task is to retrieve passages from a collection of web documents containing about 8.8 million passages.
All questions in this dataset are sampled from real and anonymized Bing queries \cite{msmarco}. 

The task provides 532k question-passage pairs for training and 6,980 pairs as development set.
Following \newcite{ance}, we report results on the MS MARCO dev set and TREC test sets from ``TREC 2019 DL" track \cite{trec2019}.
We report our results using as metrics the Mean Reciprocal Rank (MRR@10) and the Recall@1k on the dev set and the Normalized Discounted Cumulative Gain (NDCG@10) on the test set.
We use the evaluation scripts provided by TREC organizers\footnote{https://github.com/usnistgov/trec\_eval}. 

We generate synthetic questions in a similar way as described above and train the question generation model on MS MARCO.
We then use the model to generate three questions for each passage in the collection, which results in 26.4 million synthetic question-passage pairs.

\begin{table*}[t!]
\centering
\begin{tabular}{l|>{\centering\arraybackslash}p{7mm}>{\centering\arraybackslash}p{7mm}>{\centering\arraybackslash}p{7mm}>{\centering\arraybackslash}p{7mm}>{\centering\arraybackslash}p{7mm}|>{\centering\arraybackslash}p{7mm}>{\centering\arraybackslash}p{7mm}>{\centering\arraybackslash}p{7mm}>{\centering\arraybackslash}p{7mm}>{\centering\arraybackslash}p{7mm}}

& \multicolumn{5}{c|}{\textbf{Natural Questions}}      & \multicolumn{5}{c}{\textbf{SQuAD}}       \\ 
& \textbf{Top 1} & \textbf{Top 5} & \textbf{Top 10} & \textbf{Top 20 }& \textbf{Top 100} & \textbf{Top 1} & \textbf{Top 5} & \textbf{Top 10} & \textbf{Top 20} & \textbf{Top 100}  \\ \hline
\textbf{Baseline models} & & & & & & & & & & \\ 
~~~~DPR (Single)      &- & -&- & 78.4 & 85.4 & -& -& -& 63.2 & 77.2 \\
~~~~DPR \textit{ours}  & 44.6 &	68.1 & 74.5 & 79.6 & 86.2 & 25.3	& 47.3 & 56.3 & 64.4 & 78.1  \\
~~~~BM25   & - & -&- & 59.1 & 73.7  & -& -& -& 68.8 & 80.0 \\
~~~~BM25 + DPR &- & -&- & 76.6 & 83.8 & -& -& -& 71.5 & 81.3 \\
~~~~ANCE (Single)      & - & -&- & 81.9 & 87.5 & -& -& -& - & -  \\ \hline 
\textbf{Our models} & & & & & & & & & & \\ 
~~~Stage 1 & 31.6 & 59.2 & 68.1 & 74.7 & 84.7 & 22.3 & 44.6 & 54.2 & 61.9 & 75.9 \\
~~~~~ + Rnd  & 40.5 & 67.2 & 75.2 & 80.6 & 87.4 & 30.1 & 53.7 & 62.0 & 69.5 & 81.2 \\
~~~~~ + Rnd + Coarse  & 42.4 & 69.3 & 77.4 & 81.8 & 88.1 & 33.7 & 57.3 & 65.8 & 72.9 & 83.8 \\
~~~~~ + Rnd + Fine  & 42.1 & 69.4 & 77.4 & 82.1 & 88.1 & 33.4 & 57.2	& 65.5 & 72.8 & 83.7  \\
~~~~~ + Rnd + BM25  & 50.0 & 72.2 & 78.1 & 82.2 & 87.7  & 30.7 & 54.4 & 63.3 & 70.9 & 82.7  \\
~~~~~ + Rnd + Context & 51.0 & 72.4 & 77.8 & 82.1 & 88.1 & 30.6 & 53.9 & 62.7 & 69.7 & 81.8\\
~~~~~ + Rnd + Mixed   & 50.3	& 72.1	& 78.2 & 82.7 & \textbf{88.6} & 33.4 & 56.8 & 65.4 & 72.4 & 83.6  \\
\cdashline{1-11}
~~~~Ensemble (embedding) & \textbf{52.0} & \textbf{73.3 }& \textbf{78.5} & \textbf{82.8} & 88.4 & \textbf{35.2} & \textbf{58.7} & \textbf{66.9} & \textbf{73.6} & \textbf{84.2} \\ 
~~~~Ensemble (rank) & 49.0 & 72.6 & 78.3 & 82.7 & 88.5 & 34.2 & 58.3 & 66.6 & 73.4 & 83.8 \\ 
\end{tabular}
\caption{Results on Open Domain QA NQ and SQuAD retrieval tasks. \textbf{[Our models]} are trained using a two-stage neural retrieval model that uses hard negatives in both stages.}
\label{qa_results}
\end{table*}

\subsection{Implementation Details}

We use the public pre-trained uncased BERT\textsubscript{Base}\footnote{https://github.com/google-research/bert} as the initial checkpoint for our models.
We encode questions and passages into vectors of size 64 and 384 respectively.
We extract 100 hard negatives for each question and in each training iteration, we randomly pick 2 hard negatives per question to append to the training batch.
We train our models for 200 epochs using Adam with learning rate of 1e-5.
We use recall@1 on the development set as signal for early stopping.
All models are trained on a ``4x4'' slice of V3 Google Cloud TPU using batches of size 2048.

\section{Results and Discussion}

\subsection{Results on Open Domain QA Retrieval}

\begin{table*}[h!]
\begin{center}
\begin{tabular}{l|l|cc|c}
   & \textbf{Encoder} & \multicolumn{2}{c|}{\textbf{MS MARCO Dev}} & \textbf{TREC DL Test}  \\
   & \textbf{Backbone} & \textbf{MRR@10} & \textbf{Recall@1k} & \textbf{NDCG@10} \\
   \cline{1-5} 
 \textbf{Baseline models} & & & & \\
 ~~~~BM25-Anserini      & - & 18.7 & 85.7 & 49.7 \\
 ~~~~ANCE               & RoBERTa & 33.0 & 95.9 & 64.8 \\ 
 ~~~~ME-BERT            & BERT\textsubscript{Large} & 33.4 & -  & 68.7 \\
 ~~~~ME-HYBRID-E        & BERT\textsubscript{Large} & 34.3 & -  & 70.6 \\ 
 ~~~~DPR \textit{ours}  & BERT\textsubscript{Base} & 23.8 & 87.1 & 56.0 \\
 \cline{1-5} 
 \textbf{Our models~(BERT\textsubscript{Base})}    &  & & & \\
 ~~~~Stage 1            & - & 26.4 & 97.4 & 60.0 \\
 ~~~~~ + Rnd                & - & 28.3 & 97.4 & 63.6  \\
 ~~~~~ + Rnd + Coarse       & - & 30.8 & 97.6 & 67.4 \\
 ~~~~~ + Rnd + Fine         & - & \textbf{31.2} & \textbf{97.7} & 67.1 \\
 ~~~~~ + Rnd + BM25         & - & 30.0 & 97.3 & 64.6 \\
 ~~~~~ + Rnd + Context      & - & 28.9 & 97.2 & 66.3 \\
 ~~~~~ + Rnd + Mixed        & - & 30.5 & 97.6 & 66.5 \\ \cdashline{1-5} 
 ~~~~Ensemble (embedding)   & - & 30.7 & 96.2  & \textbf{67.5}  \\ 
 ~~~~Ensemble (rank)        & - & 31.1 & \textbf{97.7} & \textbf{67.5} \\ 
\end{tabular}
\end{center}
\caption{Results on MS MARCO Dev and TREC DL Test set. \textbf{[Our models]} are trained using a two-stage neural retrieval model that uses hard negatives in both stages based on BERT\textsubscript{Base}}
\label{tab:msmarco_results}
\end{table*}

The results for our system on the open domain QA retrieval tasks are shown in Table~\ref{qa_results}. 
Rows 1 to 5 show the results of the baseline systems.
The \textbf{DPR} system using the dual encoder model proposed by \newcite{dpr}\footnote{It corresponds to the \textit{Single} version in their paper that trains the model on one dataset only}.
For the sake of reproducibility, we re-implemented the DPR system as described in section \ref{sec:model}.
In contrast to ours, the DPR model does not share the question encoder and the passage encoder from the BERT model.
Instead, it uses separate encoders for each type of text.
Moreover, it does not have an additional fully connected projection layer and the loss function is different:
DPR uses batch-softmax while we use bidirectional batch-softmax.
With these modifications, and as shown in the second row, our implementation \textbf{(DPR ours)} outperforms the original DPR on both the NQ and SQuAD evaluations.
Subsequent rows show the performance of a strong sparse model \textbf{BM25} and a hybrid model, \textbf{BM25+DPR}. 
It is not surprising that the results of the sparse model on SQuAD outperform the results of the dense models in the first and second rows.
The hybrid model is formulated to first retrieve the top 2,000 passages according to BM25 and then rerank them using a ranking function that is a linear combination of the BM25 scores and DPR scores.
Finally, we show the results of \textbf{ANCE} \cite{ance}, which is also a dual encoder model and has recently shown strong results in the NQ task including top performance in terms of Top20 and Top100 accuracy.
Similar to our model, ANCE explores the use of hard negatives.
ANCE constructs hard negatives from an approximate nearest neighbor index of the corpus that is updated periodically during training.
In contrast, we generate the hard negatives once. Note that ANCE is started from the released DPR checkpoints. 

The results of our models are shown in rows~6 through 14.
Row~6 shows the results of the Stage 1 model pre-trained using synthetic data with context hard negatives; no fine tuning. 
The models in the rest of table use the same Stage 1 model but additionally they utilize different hard negatives during fine tuning. 
Row~7 shows the result of the model that uses in-batch random~(Rnd) negatives only during fine tuning.
Despite this being our initial approach, it is interesting to notice that the accuracy rates on NQ are already very close to the results of ANCE, and the accuracy rates on NQ and SQuAD outperform the ones of both BM25 and DPR. 
Rows 8 through 12 show that the models trained with different types of hard negatives during fine-tuning improve the performance over the model that does not use them (row~7).
They also outperform the results of prior works on both NQ and SQuAD. 
Specifically, when using context hard negatives, our models achieve the best Top1 and Top5 accuracy rates on NQ and get a remarkable improvement of 6.4 points and 4.3 points respectively over DPR.
The Top10/Top20/Top100 accuracy rates for the four kinds of hard negatives are all very similar.
On SQuAD, the model that uses coarse hard negatives achieves the best accuracy rates and outperforms the hybrid BM25+DPR model by 1.4 points on Top5 accuracy and 2.5 points on Top10 accuracy.
We reason that the performance difference between NQ and SQuAD is originated by how the datasets were created,
and the fact that SQuAD has much larger token overlap between questions and passages in comparison to NQ.

We also experimented with mixing all 4 hard negatives during training.
Mixing consists of using all four kinds of negatives in the pool to sample $N$ hard negatives during training (section \ref{sec:hnintraining}).
The best Top10/Top20/Top100 accuracy is achieved on NQ with this setting. 
Lastly, we show the result of the ensemble models. Notably, we found that embedding fusion further improved the Top 1 accuracy by 1.0 and 1.5 points on NQ and SQuAD, respectively. Other accuracy rates also improved except the Top 100 accuracy on NQ.

\begin{table*}[t!]
\centering
\begin{tabular}{l|>{\centering\arraybackslash}p{7mm}>{\centering\arraybackslash}p{7mm}>{\centering\arraybackslash}p{7mm}>{\centering\arraybackslash}p{7mm}>{\centering\arraybackslash}p{7mm}|>{\centering\arraybackslash}p{7mm}>{\centering\arraybackslash}p{7mm}>{\centering\arraybackslash}p{7mm}>{\centering\arraybackslash}p{7mm}>{\centering\arraybackslash}p{7mm}}
& \multicolumn{5}{c|}{ \textbf{Natural Questions}}    & \multicolumn{5}{c}{ \textbf{SQuAD}}       \\ 
& \textbf{Top 1} & \textbf{Top 5} & \textbf{Top 10} & \textbf{Top 20 }& \textbf{Top 100} & \textbf{Top 1} & \textbf{Top 5} & \textbf{Top 10} & \textbf{Top 20} & \textbf{Top 100}  \\ \hline
\textbf{Full model} & & & & & & & & & & \\ 
 ~~~~Rnd  & 40.5 & 67.2 & 75.2 & 80.6 & 87.4 & 30.1 & 53.7 & 62.0 & 69.5 & 81.2 \\
 ~~~~Rnd + Coarse  & 42.4 & 69.3 & 77.4 & 81.8 & 88.1 & 33.7 & 57.3 & 65.8 & 72.9 & 83.8 \\
 ~~~~Rnd + Fine  & 42.1 & 69.4 & 77.4 & 82.1 & 88.1 & 33.4	& 57.2	& 65.5 & 72.8 & 83.7  \\
 ~~~~Rnd + BM25  & 50.0 & 72.2 & 78.1 & 82.2 & 87.7  & 30.7 & 54.4 & 63.3 & 70.9 & 82.7  \\
 ~~~~Rnd + Context & 51.0 & 72.4 & 77.8 & 82.1 & 88.1 & 30.6 & 53.9 & 62.7 & 69.7 & 81.8\\
 ~~~~Rnd + Mixed   &  50.3	& 72.1	& 78.2 & 82.7	& 88.6 & 33.4 & 56.8 & 65.4 & 72.4 & 83.6  \\
\hline
\multicolumn{5}{l}{\textbf{No hard negative in Stage 1}}  & & & & & & \\ 
 ~~~~Rnd  & 39.0 & 65.9 & 73.9 & 79.8 & 87.1  & 28.1 & 51.3 & 59.9 & 67.2 & 80.0 \\
 ~~~~Rnd + Coarse  & 41.3 & 68.0 & 76.0 & 81.6 & 88.2  & 30.7 & 54.6 & 63.3 & 70.7 & 82.6 \\
 ~~~~Rnd + Fine  & 39.7 & 67.6 & 76.0 & 81.2 & 88.0 & 31.0 & 54.3 & 63.4 & 70.5 & 82.5 \\
 ~~~~Rnd + BM25 & 47.3 & 70.8 & 77.0 & 81.4 & 87.4 & 27.0 & 50.3 & 59.8 & 67.9 & 80.8 \\
 ~~~~Rnd + Context  & 49.4 & 71.2 & 77.2 & 81.4 & 87.6 & 28.8 & 52.4 & 61.3 & 68.5 & 80.9 \\
 ~~~~Rnd + Mixed & 49.6 & 71.9 & 77.9 & 82.6 & 88.0 & 30.1 & 54.3 & 63.0 & 70.4 & 82.7  \\ 
\hline
\multicolumn{1}{l|}{\textbf{No Stage 1}} & & & & & & & & & & \\  
 ~~~~Rnd    & 35.9 & 62.2 & 70.3 & 77.2 & 85.5 & 25.0 & 46.5 & 54.7 & 62.4 & 75.7 \\
 ~~~~Rnd + Coarse & 35.9 & 62.4	& 70.8 & 77.5 & 85.6 & 24.9 & 46.3 & 55.2 & 63.2 & 77.3  \\
 ~~~~Rnd + Fine & 36.3 & 63.1 & 71.3 & 77.8 & 85.7 & 25.5 & 46.7 & 55.3 & 63.6 & 77.5  \\
 ~~~~Rnd + BM25 & 44.6 &	68.1 & 74.5 & 79.6 & 86.2 & 25.3	& 47.3 & 56.3 & 64.4 & 78.1  \\
 ~~~~Rnd + Context & 47.6 & 68.7 & 75.4 & 79.8 &	86.4 & 26.6 & 48.0 & 56.3 & 63.8 & 76.7  \\
 ~~~~Rnd + Mixed  & 39.1 & 63.8 & 72.4 & 78.1 &	85.9 & 28.0 & 50.8 & 59.1 & 66.8 & 79.6 \\
\end{tabular}
\caption{Model Ablation Results on Open Domain QA NQ and SQuAD retrieval tasks by removing the hard negatives in stage 1 and removing stage 1 completely.}
\label{tab:ablation}
\end{table*}

\subsection{Results on MS MARCO Passage Ranking}
\label{sec:msmarcoexperiments}

Table~\ref{tab:msmarco_results} shows the results on MS MARCO Dev set and TREC DL Test set.
The top half of the table shows the results of baseline models.
The dense retrieval models, including ANCE, ME-BERT and ME-HYBRID-E \cite{luan2020sparse}, significantly outperform BM25-Anserini\cite{anserini} with parameters k1=0.82, b=0.68.
ME-BERT is a model in which every passage is represented by multiple vectors from BERT.
ME-HYBRID-E is a hybrid model of ME-BERT and BM25-Anserini which linearly combines sparse and dense scores using a single trainable weight. 
Note that ANCE is initialized with RoBERTa\textsubscript{Base} and ME-BERT and ME-HYBRID-E are initialized with BERT\textsubscript{Large}.
As reference for the performance gains from our improved negative contrast, we also include as baseline our implementation of DPR, which is based on BERT\textsubscript{Base}. 

The bottom half of the table shows the results of our model.
Our Stage 1 model is trained with coarse hard negatives as the mapping between passages to documents is not available in this case. 
The Stage 1 model outperforms our DPR baseline on both sets and all three metrics.
This indicates that the DPR model can benefit from more training examples.
When we fine tune the Stage 1 model with in-batch (Rnd) negatives, the model improves by 1.9 points in MRR@10 on dev set and 3.6 points in NDCG@10 on test set.
When using different hard negatives, the models outperform the model that does not use them.
The model that uses fine hard negatives achieves the best MRR@10 among all four types of hard negatives. 
The recall@1k for the different types of hard negatives are very similar.
The best NDCG@10 is achieved when the model is trained with coarse hard negatives. 
Both ensemble models perform similarly on these metrics. 

\subsection{Model Ablations}
We conduct ablation experiments in order to understand the contribution of each component in our model.
We show the results on the open-domain QA datasets in Table~\ref{tab:ablation}.
Rows 1 to 6 present the performance of the models using the full two-stages training reported in the second half of Table~\ref{qa_results}.
We first remove the hard negatives in Stage 1 but keep them in the fine tuning stage.
As shown in rows 7-12, the accuracy rates drop across all settings on both NQ and SQuAD.
This shows that the context hard negatives benefit the training with synthetic data and that using hard negatives in both stages is the best performing option.
We go further and remove the Stage 1 training altogether.
In this way we are fine tuning directly on the BERT checkpoint.
Rows 13-18 show that the performance drops on both datasets significantly and points to the fact that using synthetic data to pre-train the system is highly effective. 

\begin{table*}[h!]
\centering
\begin{small}
\begin{tabular}{ |l|p{13cm}| } 
 \hline
 Question & Who sings the song Never Be the Same  \\
 Answer & Camila Cabello \\
 Gold & Never Be the Same (Camila Cabello song) ``Never Be the Same'' is a song by Cuban-American singer Camila Cabello from her debut studio album, ``Camila'' (2018). The song was written by Cabello, Noonie Bao and Sasha Yatchenko ...... \\ \hline \hline
 Random & American Civil War and Cuba's Ten Years' War, U.S. businessmen began monopolizing the devalued sugar markets in Cuba. In 1894, 90\% of Cuba's total exports went to the United States... \\ \hline
 Coarse & and his group the Bob-cats. In 2008 Crosby's rendition of the song appeared as part of the soundtrack of ``Fallout 3''. The song made a repeat appearance in ``Fallout 4'' in 2015. Happy Times (song) ``Happy Times'' is a jazz ballad written by American lyricist Sylvia Fine ...... \\ \hline
 
 Fine & Sisters (song) "Sisters" is a popular song written by Irving Berlin in 1954, best known from the 1954 movie ``White Christmas''. Both parts were sung by Rosemary Clooney (who served as Vera-Ellen's singing vocal dub for this song, while Trudy Stevens dubbed Vera-Ellen's ...... \\ \hline
 
 BM25 & release of the album. The song has been certified Gold by the British Phonographic Industry (BPI). An accompanying but unofficial music video for ``Never Be the Same'' was released on Cabello's personal YouTube channel on December 29, 2017...... \\ \hline
 
 Context & ``Never Be the Same'' has been described as a ``dark'' pop ballad. A ``NME'' writer described it as ``bombastic'' electro. The upbeat track features Cabello singing falsetto in the pre-chorus. According to sheet music published by Sony/ATV Music Publishing on Musicnotes.com, ......\\ 

 \hline
\end{tabular}
\end{small}
\caption{Examples of hard negatives}
\label{tab:example}
\end{table*}

\subsection{Hard Negatives Examples}
\label{sec:example}

Table~\ref{tab:example} shows examples of the four types of hard negatives and a random negative, which was selected among the passages in one of the training batches from NQ {\footnote{Examples on MS MARCO data set can be found in the Appendix.}}, for reference.
Given a question and its gold passage, the coarse hard negative passage is on topic, about a song, but not about the song mentioned in the question.
The fine hard negative passage describes a different song from the one in the question but it mentions the singer of the song discussed.
This singer-song relationship is semantically close to the similar relationship observed in the gold passage.
Both the BM25 hard negative passage and the context hard negative passage mention the song in the question and they are semantically closer to the gold passage in comparison to the coarse and fine hard negatives.

\section{Conclusion}
We presented four strategies to sample hard negatives in the context of automatic question answering.
Our experiments on Natural Questions, SQuAD and MS MARCO show that our strategies are highly effective to improve the contrast between the passage that answers a question and all the other passages.
Our models improved on the previous state-of-the-art for Open Domain QA by 0.8--2.9 points.
We trained BERT-based dual encoder models using a two-stage approach.
We demonstrated the positive impact of the hard negatives in both stages.
Our changes to the model architecture already showed improvement over the baseline.
We further improved the models with the use of synthetic data in the pre-training stage.
The use of hard negatives at this stage showed considerable gains over the baseline.
When we used four different types of hard negatives in the fine-tuning stage, we achieved state-of-the-art performance in passage retrieval.
Mixing the different types of hard negatives proved to be the most effective strategy.
Still, we achieved bigger gains when we used emsembling to combine the benefits of all the different types of hard negatives.
Our results encourage us to keep exploring this area and investigate similar contrastive mechanisms to improve reading comprehension in end-to-end question answering systems.

\section*{Acknowledgement}
The authors would like to thank Vladimir Magay for valuable discussion, Daniel Cer for reviewing the manuscript and Fangxiaoyu Feng for advice on inference for indexing.

\bibliography{qa}
\bibliographystyle{acl_natbib}

\appendix
\section{Synthetic Data Examples}
Table~\ref{tab:synthetic_example} shows several examples of synthetic questions. The first two are from open-domain QA and the last two are from MS MARCO. As shown in the table, even though they are synthetic questions, they are of high quality. In addition, we can see that questions in these two tasks have different styles.

\begin{table*}[t!]
\centering
\begin{small}
\begin{tabular}{ l|l } 
\hline
 \textbf{Passage} & \textbf{Synthetic Questions} \\ \hline
 \multirow{3}{*}{\parbox{8cm}{North Park Secondary School is a public high school located at the major intersection of Williams Parkway and North Park Drive in Brampton, Ontario, Canada. It was founded in 1978, making it one of the oldest high schools in the area. North Park is best known for being one of three high schools in Brampton to offer the IBT program, a program using business and technology to enrich the learning of its students. Students in the IBT program are often required to bring a device such as a laptop to guide through courses by filing notes}} & why do students go to north park school \\
& what is the type of school north park high school \\
& where is north park secondary school in brampton ontario \\ [60pt]

\hline
 
 \multirow{3}{*}{\parbox{8cm}{
age 18 and over, there were 93.1 males. The median income for a household in the CDP was \$43,125, and the median income for a family was \$45,327. Males had a median income of \$36,524 versus \$29,861 for females. The per capita income for the CDP was \$19,670. About 9.7\% of families and 9.9\% of the population were below the poverty line, including 14.5\% of those under age 18 and 8.7\% of those age 65 or over. In the state legislature, Valley Springs is in , and . Federally, Valley Springs is in . Valley Springs, California Valley Springs (formerly,}} & what is the poverty line in valley springs ca \\
& what is the median income in valley springs ca \\
& what is the largest city in the central valley of california \\ [60pt]
\hline

 \multirow{3}{*}{\parbox{8cm}{Start recording at any time during a conference call. Control as you record by pausing and resuming recording. Recording can be initiated by any touch-tone phone. Playback toll-free via phone access, start, stop, rewind and fast forward at your control using touch-tone commands on the phone keypad.}} & are tap phones recording \\
 & how to record a conference call \\
 & can you see what you record on your phone \\ [25pt]
  \hline

\multirow{3}{*}{\parbox{8cm}{Updated PANDAS signs and symptoms (1) Pediatric onset. The first symptoms of PANDAS are most likely to occur between 5 and 7 years of age. Symptoms can occur as early as 18 months of age or as late as 10 years of age. If the first clinically recognized episode is detected after the age of 10, it is unlikely true initial episode, but the recurrent one.}} & child pandas symptoms \\
& age of onset of pandas \\
& what age can you be affected by pandas \\ [35pt]

\hline

\end{tabular}
\end{small}
\caption{Examples of Synthetic Data}
\label{tab:synthetic_example}
\end{table*}

\section{MS MARCO Hard Negative Examples}
Table~\ref{tab:msmarco_example} shows examples of four types of hard negatives plus a random negative in one of the training batches from MS MARCO dataset.

\begin{table*}[h!]
\centering
\begin{small}
\begin{tabular}{ |l|p{13cm}| } 
 \hline
 Question & Genetic Predispositions definition psychology  \\
 Gold &  A genetic predisposition is a genetic effect which influences the phenotype of an organism but which can be modified by the environmental conditions. Genetic testing is able to identify individuals who are genetically predisposed to certain health problems.redisposition is the capacity we are born with to learn things such as language and concept of self. Negative environmental influences may block the predisposition (ability) we have to do some things. \\ \hline \hline
 Random & They're loaded with nutrients, called antioxidants, that are good for you. Add more fruits and vegetables of any kind to your diet. It'll help your health. Some foods are higher in antioxidants than others, though. The three major antioxidant vitamins are beta-carotene, vitamin C, and vitamin E.  \\ \hline
 
 Coarse & Prevention of Musculoskeletal Disorders in the Workplace. Musculoskeletal disorders (MSDs) affect the muscles, nerves and tendons. Work related MSDs (including those of the neck, upper extremities and low back) are one of the leading causes of lost workday injury and illness. \\ \hline
 
 Fine &  Mycoplasma pneumoniae (M. pneumoniae) is an atypical bacterium (the singular form of bacteria) that causes lung infection. It is a common cause of community-acquired pneumonia (lung infections developed outside of a hospital).M. pneumoniae infections are sometimes referred to as walking pneumonia..n general, M. pneumoniae infection is a mild illness that is most common in young adults and school-aged children. The most common type of illness caused by these bacteria, especially in children, is tracheobronchitis, commonly called a chest cold. \\ \hline
 
 BM25 &  There is definitely a genetic predisposition to arterial disease and the risk factors that cause it.There have been certain genetic abnormalities that have been identified.here is definitely a genetic predisposition to arterial disease and the risk factors that cause it. \\ \hline
 
 Context & ... are at risk for loss of health insurance if they are discovered to have genetic predispositions for health problems. The national center for genome resources found that 85 percent of those polled think employers should not have access to information about their employees genetic conditions risks or predispositions. 2 the us federal government has so far taken only limited measures against discrimination based on genetic testing... \\ 

 \hline
\end{tabular}
\end{small}
\caption{Examples of hard negatives on MS MARCO dataset.}
\label{tab:msmarco_example}
\end{table*}

\end{document}